\documentclass{article}
\usepackage{arxiv}
\usepackage{listings}

\usepackage{soul}
\usepackage{url}
\usepackage[small]{caption}
\usepackage{graphicx}
\usepackage{amsmath}
\usepackage{amsthm}
\usepackage{algorithm}
\usepackage{graphicx}
\usepackage{amsmath}
\usepackage{amsfonts}
\usepackage{booktabs}
\usepackage{multirow}
\usepackage{xcolor}
\usepackage{comment}
\usepackage{breqn}
\usepackage{algpseudocode}

\newtheorem{definition}{Definition}
\newtheorem{property}{Property}

\title{Analogical relevance index}

\author{{Suryani Lim}\\
Federation University, Churchill, Australia\\
\texttt{suryani.lim@federation.edu.au}
\And {Henri Prade}\\
IRIT, University of Toulouse, France\\
\texttt{henri.prade@irit.fr}
\And {Gilles Richard}\\
IRIT, University of Toulouse, France\\
\texttt{gilles.richard@irit.fr}
}

\algnewcommand\algorithmicforeach{\textbf{for each}}
\algdef{S}[FOR]{ForEach}[1]{\algorithmicforeach\ #1\ \algorithmicdo}

\begin{document}

\maketitle

\begin{abstract}
Focusing on the most significant features of a dataset is useful both in machine learning (ML) and data mining. In ML, it can lead to a higher accuracy, a faster learning process, and ultimately a simpler and more understandable model. In data mining, identifying significant features is essential  not only for gaining a better understanding of the data but also for visualisation. In this paper, we demonstrate a new way of identifying significant features inspired by analogical proportions. Such a proportion is of the form of "a is to b as c is to d", comparing two pairs of items $(a,b)$ and $(c,d)$ in terms of similarities and dissimilarities. In a classification context, if the similarities/dissimilarities between $a$ and $b$ correlate with the fact that $a$ and $b$ have different labels, this knowledge can be transferred to $c$ and $d$, inferring that $c$ and $d$ also have different labels. From a feature selection perspective, observing a huge number of such pairs $(a,b)$ where $a$ and $b$ have different labels provides a hint about the importance of the features where $a$ and $b$ differ. Following this idea, we introduce the Analogical Relevance Index ($ARI$), a new statistical test  of the significance of a given feature with respect to the label. 
$ARI$ is a filter-based method. Filter-based methods are ML-agnostic but generally unable to handle feature redundancy. However,  $ARI$ can detect feature redundancy. Our experiments show that $ARI$ is effective and outperforms well-known methods on a variety of artificial and some real datasets.
\end{abstract}
\section{Introduction}\label{intro}
Representing real-life data often implies the use of many features, leading to an item being represented by a high-dimension vector. Note that in this paper, we use indistinctly the word 'feature' or 'attribute' or even 'variable'.
Managing all these features contributes to the need for high computational power in terms of space and time, and could slow down the construction of a predictive model and degrade the predictive accuracy of the constructed model.
Two options are then available:
\begin{enumerate}
    \item Dimension reduction by projecting the initial dataset of high dimensionality to a new feature space with lower dimensionality. However, in the new space, there is no guarantee that all the features are relevant for predicting the label.
    Principal Component Analysis (PCA) \cite{PCA2013} is one of the most popular techniques.
    \item Feature selection, on the other hand, directly selects a strict subset from all available features.
    In a classification task, it is often the case that only a subset of features is {\it relevant} i.e. nonredundant and informative in determining the label.
    Unfortunately, in the real world, without a deep knowledge of the problem domain, relevant features are rarely a priori knowledge. 
\end{enumerate}

Feature selection methods can be classified into diverse categories; for more information, please see \cite{LietalACM2018} for an exhaustive investigation). Generally, they can be classified as:
\begin{itemize}
    \item Filter methods, where features are independently evaluated and associated to a relevance score. They are independent of any ML algorithm (ML agnostic) as the filtering is done by observing the data only. Usually, these methods cannot detect redundancy.
    \item Wrapper methods, which initially evaluate all features for an ML algorithm. Then, select a strict subset of the features, then retrain and retest.
    The whole process is embedded in a loop on the power set of the features. 
    The features belonging to the subset that gives the highest accuracy are considered relevant. It is effectively, an exhaustive search, so it is highly resource-consuming. 
    \item Embedded methods are a trade-off between filter and wrapper methods where feature selection is embedded into the training process of an ML algorithm. They generally provide good results but the selected features are tailored to the chosen ML algorithm. This implies that the selected features may not be the most effective for other algorithms.
\end{itemize}

Filter-based methods generally proceed in 2 steps:
\begin{itemize}
    \item Score each feature individually: 
    the higher the score, the more relevant the feature.
    \item Eliminate all features whose scores are below a given threshold. 
\end{itemize}
In this paper, we propose Analogical Relevant Index ($ARI$), a new filter-based method for feature selection: as such $ARI$ is ML agnostic. $ARI$ is based on a statistical test and is inspired by the concept of analogical proportion. An analogical proportion is a statement involving 4 items: "a is to b as c is to d", and it is often denoted as $a:b::c:d$. Its semantics is that $a$ differs from $b$ as $c$ differs from $d$, where $a,b,c,d$ are items represented as vectors of the same dimensions.
Because pairs of items are used as the initial observations as in the case of analogical proportions, hence the word "Analogical" in ARI.

Our paper is structured as follows: in Section \ref{related}, we provide an overview of related work.
In Section \ref{context}, we provide the context and notations.
Section \ref{ARI} develops the complete formal framework necessary to define $ARI$. We also investigate the theoretical properties of such an index.
In Section \ref{expe}, we define our experimental context and exhibit the results from diverse experiments. We also compare $ARI$ to other filter-based approaches. Section~\ref{discussion} discusses the stability and limitation of $ARI$.
Finally, in Section \ref{conc}, we provide concluding remarks and tracks for future investigations.

All our code is freely available on https://github.com/gillesirit/ARI.
\section{Related work}\label{related}
The research landscape in feature selection is overcrowded and it is challenging to select specific papers strictly related to our work.
A good starting point could be \cite{Guyon2004ResultAO} challenge from NIPS2003. Despite being quite old, many interesting things could be learned. Since then, the ML revolution of deep learning occurred and brought new highlights. We adopt a more restrictive approach in this paper by focusing on filter-based methods, i.e. methods that are independent of ML algorithms, and with a limited number of available features. We have to admit that this could drastically reduce the range of applications of our proposed method.
In that perspective,  one of the most related works we can cite is that of \cite{KeaSmyICCBR2020}. Although their work is not focused on feature selection, they highlight the issues of case-based or analogy-based reasoning.

\section{Context and notations}\label{context}
In this paper, we deal with categorical features and categorical output (label). 
A typical example of a categorical feature is 'color' with values such as $\{R, G, B\}$. 
Obviously, we can replace the letters by numbers $\{1,2,3\}$ but any arithmetic operations on the numbers such as $1-3$ are meaningless. 
Categorical data can be defined as follows:
\begin{itemize}
    \item An instance $a$ belongs to a Cartesian product $X$ of finite sets $X_i$: $X = X_1 \times \ldots \times X_n$.
    \item We consider feature names as their index so that the set of features is $\mathcal{A}=\{1,\ldots, n\}$.
    \item Each feature $i$ takes its values from  $X_i$.
    \item An instance $a$ is represented by a vector $(a_1,\ldots,a_n)$ of $n$ feature values $a_i \in X_i$. 
    \item Every instance $a$ is associated with a unique element (its label) belonging to a finite set $Y$.
\end{itemize}
Given a label $l \in Y$,  class $l$ is the subset of $X$ whose elements have label $l$.
An observation is an instance $a$ with its associated label $l$.  
We assume $m$ observations i.e. a finite set $S \subseteq X$ of instances $a$ where we have the corresponding label $cl(a) \in Y$. We also assume that there is no missing value in $S$ so that for every instance, every feature has a corresponding value.

In the particular case where $X=\mathbb{B}^n$ and $Y=\mathbb{B}=\{0,1\}$, it is common to consider the elements of $X$ with label $1$ as a concept $c$. 
As such, a concept $c$ can be defined via a Boolean function $f$ or via a Boolean formula $F$:
$$c = \{ x \in X | f(x)=1 \} \mbox{ or } c = \{ x \in X | F(x)  holds \}$$
For Boolean concepts and features, a definition of relevance or irrelevance has been given in \cite{AlmDieAAAI1991}. However, as we deal with categorical data, which is broader than Boolean, the definition of \cite{AlmDieAAAI1991} does not apply. We provide in the following section a general definition that is applicable to any type of categorical feature.

\section{Analogical Relevance Index}\label{ARI}
Here, we first provide the concepts and definitions needed to define Analogical Relevance Index $ARI$. Then we investigate its properties and finally discuss the impact of the curse of dimensionality.
\subsection{Main concepts and definitions}
To identify relevant or irrelevant features, our main idea is to compare pairs of instances in $S$, $(a,b)$ and $(c,d)$,  looking for similarities/dissimilarities and observing the labeling behavior (having the same or different labels). 
In fact, comparing the 2 pairs amounts to grouping the 2 pairs into an analogical proportion $a:b::c:d$ (see \cite{PraRicLU2013} for more information)
and to check if an analogical proportion is still valid for their labels.
\begin{definition} 
Given  a pair $(a,b) \in X \times X$ of vectors, their agreement set $Ag(a,b)$ is:
$$Ag(a,b)=_{def}\{i \in [1,n] | a_i =b_i \}$$

Their disagreement set $Dag(a,b)$ is:
$$Dag(a,b)=_{def}\{i \in [1,n] | a_i \neq b_i\}$$
\end{definition}
As usual the Hamming distance $H(a,b)$ between $a$ and $b$ is  $|Dag(a,b)|$. 
For instance, with dimension $n=5$:
\begin{itemize}
    \item $a=(A,yellow,45,male,0)$ 
    \item $b=(A,red,51,female,0)$
    \item $Ag(a,b)=\{1,5\}, Dag(a,b)=\{2,3,4\}$
    \item $H(a,b)=3$.
\end{itemize}
With our notations, some immediate properties are:
\begin{property} $Ag(a,b) \cup Dag(a,b) =\mathcal{A}=\{1,\ldots, n\}, \\ |Ag(a,b)|+|Dag(a,b)| = n$ \end{property}
\begin{property}
$a:b::c:d \implies Ag(a,b)= Ag(c,d), Dag(a,b)= Dag(c,d), H(a,b)=H(c,d)$.

The reverse implication is not valid as can be seen with 
\begin{itemize}
    \item $a=(A,yellow,45,male,0)$ 
    \item $b=(A,red,51,female,0)$
    \item $c=b$, $d=a$
\end{itemize}
$a:b:c:d$ does not hold but still we have
$Ag(a,b)= Ag(c,d), Dag(a,b)= Dag(c,d), H(a,b)=H(c,d)=3$.
\end{property}
We start by observing the available datasets,
$S \subseteq X=X_1 \times \ldots \times X_n$, 
the impact of a feature change on the class change.
Let be $i \in \mathcal{A}$ and let us consider the set $Dif(i)^S$ of pairs of elements among the observable data which exactly disagree on feature $i$:
$$Dif(i)^S=_{def} \{(a,b) \in S \times S | Dag(a,b)= \{i\}\}$$
For such pair of elements $(a,b)$, the agreement set is just $\mathcal{A} \setminus \{i\}$ and necessarily $a \neq b$.
Let us discuss the case where $Dif(i)^S=\emptyset$, which can be the result of two different situations:
\begin{enumerate}
    \item Either all elements in $S$ agree on feature $i$ i.e. $i$ has zero-variance on $S$. This can be checked by considering the following set:
    $$Val(i)^S =_{def} \{x \in X_i | \exists a \in S \mbox{ such that } a_i=x\}$$
    $Val(i)^S$ is the set of values taken by feature $i$ on sample $S$.
    $Val(i)^S$ cannot be empty due to our assumption on $S$, but it could be the case that  $|Val(i)^S|=1$. In that case, $i$ would be eliminated via a low variance-based feature selection method \cite{LietalACM2018}.  
    Note that $|Val(i)^S|=1 \implies Dif(i)^S=\emptyset$ but the reverse implication does not hold.
    \item Or it is impossible to differ only on feature $i$. In other words, differing on $i$ implies differing on at least another feature. In that case, we might have  dependency or redundancy between variables. For instance, if we observe on each item in $S$ that the value of feature $i$ is always equal to the value of another feature $j$, then both $Dif(i)^S$ and $Dif(j)^S$ will be empty: $j$ (or $i$) is redundant. 
\end{enumerate}
In the following definition, we assume $Dif(i)^S \neq \emptyset$.
\begin{definition} Relevance/Irrelevance are defined as follows:

Feature $i$ is relevant w.r.t $S$ iff :
$$\exists a, b \in S, [(a, b) \in Dif(i)^S \wedge cl(a) \neq cl(b)]$$

Feature $i$ is irrelevant w.r.t $S$ iff:
$$\forall a, b \in S, [(a, b) \in Dif(i)^S \implies cl(a)=cl(b)]$$
\end{definition}
At this stage,  relevance/irrelevance are dual concepts related to a sample set $S$.
\begin{property}  We have the following properties of monotony:
\begin{itemize}
    \item $\forall i \in \mathcal{A}, [S \subseteq T \implies Dif(i)^S \subseteq Dif(i)^T]$
    \item $\forall S \subseteq T, [i \mbox{ relevant w.r.t } S  \implies i \mbox{ relevant w.r.t } T]$
    \item $\forall S \subseteq T, [i \mbox{ irrelevant w.r.t } T  \implies i \mbox{ irrelevant w.r.t } S]$
\end{itemize}
\end{property}\noindent


Generally, for a sample $S$, $Dif(i)^S$ may contain pairs $(a,b)$ with $cl(a)\neq cl(b)$ and pairs $(a',b')$ with
$cl(a')= cl(b')$. So it makes sense to measure to what extent a feature is relevant w.r.t. $S$.
Let us define $Dif\_Eq(i)^S$ as a subset of $Dif(i)^S$: $$Dif\_Eq(i)^S=_{def}\{(a,b) \in Dif(i)^S|cl(a)=cl(b)\}$$
As an obvious consequence, we have:
\begin{property} 
$$\forall i \in \mathcal{A}, [S \subseteq T \implies Dif\_Eq(i)^S \subseteq Dif\_Eq(i)^T]$$
\end{property}
We then define the following Analogical Relevance Index w.r.t. $S$ denoted $ARI(i)^S$.
\begin{definition}
The Analogical Relevance Index $ARI(i)^S$ is defined as: 

if $|Val(i)^S|=1$ then $ARI(i)^S=0$

\quad\quad\quad else if  $Dif(i)^S=\emptyset$ then $ARI(i)^S=2$ 

\quad\quad\quad\quad\quad \quad\quad\quad \quad \quad \quad else
$$ARI(i)^S=\frac{|Dif(i)^S \setminus Dif\_Eq(i)^S |}{|Dif(i)^S|}$$
\end{definition}

\begin{itemize}
\item If $ARI(i)^S$ is $0$ (or close to $0$), we can consider feature $i$ has no impact w.r.t. $S$ on the label: knowing the value of feature $i$ brings no relevant information about the label of an element. 
\item On the opposite side, if $ARI(i)^S$ is $1$ (or close to $1$), it means each time feature $i$ changes, the label changes. This feature has a huge impact on the label. 
\item If $ARI(i)^S=2$, it means that we have redundancy (or dependency).
\end{itemize}
If there is no index $ARI(i)^S$ close to $0$ or close to $1$, it simply means that there is no specific feature more relevant than another one. For instance, when the label is decided by the sum of the features such as $a_1+\ldots + a_n = k$, it is likely (depending on the sample $S$) that all features are relevant to some extent. 

Ideally, having the entire $X$ at our disposal, we could compute $ARI(i)^X$. Because we generally have no access to the whole universe $X$ (especially with high dimensions), this value $ARI(i)^X$ should be considered a theoretical one. Nevertheless, we can investigate how $ARI(i)^S$ evolves when the size of $S$ increases toward the size of $X$.
Because $|Dif\_Eq(i)^S|$ and $|Dif(i)^S|$ are both increasing functions of $S$, at this stage, we cannot conclude the behavior of $ARI(i)^S$. Nevertheless, let us equip $X$ with a uniform probability measure denoted as $P$ (and $X \times X$ with the product still denoted as $P$). A sample $S_m$ of size $m$ can be considered as the result of a random process involving $m$ i.i.d. random variables $\{x^{(i)} | i \in [1,m] \}$, $x^{(i)}$ taking its values in $X_i$.
Then let us denote $ARI(i)^m=_{def}ARI(i)^{S_m}$ (with similar convention for $Dif(i)^m$ and $Dif\_Eq(i)^m$) for a sample of size $m$:
$$ARI(i)^m= \frac{|Dif(i)^m \setminus Dif\_Eq(i)^m|}{|Dif(i)^m|}$$
Then $ARI(i)^m$ is a random variable and we have the following result:
\begin{property}
For any feature $i$, $ARI(i)^m$ converges almost surely towards $ARI(i)^X$ as $m$ tends
to $|X|$:
$$ P(lim_{m \rightarrow |X|}ARI(i)^m = ARI(i)^X)=1$$
\end{property}
\begin{proof}
Due to the monotony of $Dif(i)^S$  and $Dif\_Eq(i)(i)^S$ w.r.t. $|S|$, we have: 
$$\frac{|Dif(i)^m \setminus Dif\_Eq(i)^m|}{|Dif(i)^X|} \leq ARI(i)^m \leq \frac{|Dif(i)^X \setminus Dif\_Eq(i)^m||}{|Dif(i)^m|}$$
It is then enough to prove the property for both the upper bound and the lower bound of $ARI(i)^m$.
Let us start with $\frac{|Dif\_Eq(i)^m|}{|Dif(i)^X|}$ and show that :
$$P(lim_{m \rightarrow |X|}\frac{|Dif\_Eq(i)^m|}{|Dif(i)^X|}= ARI(i)^X)=1$$
Because we deal with subsets of $X \times X$, let us denote $pr_1$ and $pr_2$ the 2 corresponding projections.
Then $$Dif\_Eq(i)^X \setminus Dif\_Eq(i)^m = \{(a,b) \in Dif\_Eq(i)^X| (a,b) \notin Dif\_Eq(i)^m\} \subseteq $$
$$\{(a,b) \in Dif\_Eq(i)^X| a \notin pr_1(Dif\_Eq(i)^m) \vee b \notin pr_2(Dif\_Eq(i)^m) \} $$
$$= \{(a,b) \in Dif\_Eq(i)^X| a \notin S^m\} \cup \{(a,b) \in Dif\_Eq(i)^X| b \notin S^m \}$$
Then $P(Dif\_Eq(i)^X \setminus Dif\_Eq(i)^m) \leq$ $$P(\{(a,b) \in Dif\_Eq(i)^X| a \notin S^m\})+P(\{(a,b) \in Dif\_Eq(i)^X| b \notin S^m\})$$
$$\leq 2 \times (|X|-m)$$
This goes to $0$ when $m$ goes to $|X|$.
A similar reasoning applies to the upper bound.
\end{proof}
The case that $P$ is not uniform has to be investigated.

\section{Experiments and results}\label{expe}

We compare $ARI$ with three other well-established methods suitable for categorical features and categorical labels:
 chi-square \cite{chi-square}, mutual information ($MI$) \cite{Shannon1948} and $Relief$ family algorithms \cite{urbanowicz2017}.
\subsection{Experimental settings}
The comparison was conducted under two experimental settings:
\begin{itemize}
    \item \textbf{Experiment setting 1} uses 8 synthetic Boolean datasets, where there is a priori knowledge about what the relevant features are. Each synthetic dataset is built with a specific function defining the label (i.e. the last column of the dataset):
    \begin{enumerate}
        \item $g_1: (x1 \neq 0 \wedge (x2 \neq 0 \vee x3 \neq 0))$
        \item $g_2: xor(x1,x2)$ where $xor(x1,x2)=(x1 \neq x2)$.
        \item $g_3: xor(xor(x1,x2),x3)$
        \item $g_4: \Sigma_1^{n}x_i=3$.
        \item $g_5: ((x_1 \neq 0) or (x_2 \neq 0) or (x_3 \neq 0)) and ((x_4 \neq 0) or (x_5=0) or (x_6 \neq 0))$, 
        \item $g_6: (x1\neq 0 \wedge ((x2 \neq 0) \vee  (x3=0)))$
        \item $g_7: \Sigma_1^{3} x_i=2$
        \item $g_8$ where the output is $1$ iff $x_1 \ldots x_{10}$ is a prime number. $g_8$ is then a particular case where relevant/irrelevant features are not known.
    \end{enumerate}
To check the effectiveness of a given method, we compare the scores of features from each  method against the known relevant features. The best method is the one whose relevant features have higher scores than irrelevant features. The best method is the one that provides high scores for known relevant features, and very low scores for irrelevant features.
    
    \item \textbf{Experiment setting 2} uses real data: we have no a priori knowledge of what the  relevant features are. The seven  datasets are from UCI: 
    \begin{enumerate}
        \item Monks1 (size: 432, dimension: 6)
        \item Monks2 (size: 601, dimension: 6)
        \item Monks3 (size: 554, dimension: 6)
        \item HIV (link:HIV-1+protease+cleavage) (size: 746, dimension 8)
        \item Primary-Tumor (size: 132, dimension: 17)
    \end{enumerate}
To estimate the effectiveness of a method, we train and test a given ML algorithm using only the relevant features suggested by each method. The best method is the one that gives the highest accuracy.
    
\end{itemize}

\subsection{Protocol}
The experiments were run using the following protocol: first, given a sample size $k$, we compute the feature relevance scores 10 times for various samples $S$ of the same size $k$. Then, we average the relevance scores and normalize the scores between $0$ and $1$. Normalization is necessary as relevance scores from each method have a different range of values. 
For Experiment setting 1, we use sample $S$ of size $k$, with $k=100, 200, 300, 500, 1000, 10000$. For Experiment setting 2, because we have far less data, we experiment with $\frac{1}{3}$ of the whole dataset.
The logic of the protocol is given in Algorithm~\ref{alg:protocol}.
\begin{algorithm}
\caption{The logic of the experiment protocol}\label{alg:protocol}
\begin{algorithmic}[1]
 
\State $relevance\_scores \gets 0$
\State $test \gets 1$

\While{$test \leq 10$}
    \State $relevance\_scores \gets relevance\_scores + feature\_selection(S)$
    \State $test \gets test + 1$
    
\EndWhile
\State $avg\_relevance\_scores = mean(relevance\_scores)$
\State $norm\_relevance\_scores = normalize(avg\_relevance\_scores)$

\end{algorithmic}
\end{algorithm}

\subsection{Selecting relevant features on synthetic data: results}
In this experiment, we want to check if $ARI$ can identify relevant features from synthetic datasets. We compare the results with three other methods: chi-squared ($\chi^2$), mutual information ($MI$) and $Relief$. The results are given in Table \ref{arr-chi-mut-artif}, and they are averaged over 10 runs, where the data of each run is made up of a third of the data set (randomly selected).
For each function, relevant features are underlined. As a consequence, a simple reading of the table should be that, when a feature is not underlined, the corresponding score should be zero or very close to zero.
\begin{table}[!ht]
\addtolength{\tabcolsep}{-1.5pt}
    \centering
    \tiny
    \begin{tabular}{c|c|c|c|c|c|c|c|c|c|c}
    $f$ & $a_1$ & $a_2$& $a_3$& $a_4$& $a_5$&$a_6$ &$a_7$ & $a_8$& $a_9$ & $a_{10}$\\
    $g_1$ & \underline{$a_1$} & \underline{$a_2$} & \underline{$a_3$} & $a_4$& $a_5$&$a_6$ &$a_7$ & $a_8$& $a_9$ & $a_{10}$\\
     $ARI$ & 0.61 & 0.19 & 0.19 & 0.0 & 0.0 & 0.0 & 0.0 & 0.0 & 0.0 & 0.0\\
        $\chi^2$ & 0.82 & 0.08 & 0.1 & 0.0 & 0.0 & 0.0 & 0.0 & 0.0 & 0.0 & 0.0\\
        $MI$ & 0.74 & 0.05 & 0.07 & 0.01 & 0.01 & 0.01 & 0.02 & 0.03 & 0.02 & 0.01\\
        $Relief$ & 0.54 & 0.13 & 0.18 & 0.02 & 0.01 & 0.01 & 0.03 & 0.02 & 0.02 & 0.04\\
    $g_2$ & \underline{$a_1$} & \underline{$a_2$} & $a_3$& $a_4$& $a_5$&$a_6$ &$a_7$ & $a_8$& $a_9$ & 4 $a_{10}$\\
        $ARI$ & 0.5 & 0.5 & 0.0 & 0.0 & 0.0 & 0.0 & 0.0 & 0.0 & 0.0 & 0.0\\
        $\chi^2$ & 0.16 & 0.0 & 0.05 & 0.03 & 0.04 & 0.0 & 0.49 & 0.0 & 0.12 & 0.09\\
        $MI$ & 0.05 & 0.08 & 0.15 & 0.04 & 0.09 & 0.02 & 0.11 & 0.07 & 0.08 & 0.1\\
        $Relief$ & 0.44 & 0.46 & 0.01 & 0.02 & 0.01 & 0.02 & 0.01 & 0.01 & 0.01 & 0.01\\
    $g_3$ & \underline{$a_1$} & \underline{$a_2$} & \underline{$a_3$} & $a_4$& $a_5$&$a_6$ &$a_7$ & $a_8$& $a_9$ & $a_{10}$\\
        $ARI$ & 0.33 & 0.33 & 0.33 & 0.0 & 0.0 & 0.0 & 0.0 & 0.0 & 0.0 & 0.0\\
        $\chi^2$ & 0.03 & 0.04 & 0.31 & 0.13 & 0.02 & 0.08 & 0.23 & 0.04 & 0.01 & 0.11\\
        $MI$ & 0.1 & 0.18 & 0.0 & 0.06 & 0.08 & 0.12 & 0.08 & 0.13 & 0.09 & 0.02\\
        $Relief$ & 0.29 & 0.34 & 0.34 & 0.0 & 0.0 & 0.0 & 0.0 & 0.0 & 0.0 & 0.01\\
    $g_4$ & \underline{$a_1$} & \underline{$a_2$} & \underline{$a_3$} &\underline{$a_4$} & \underline{$a_5$} & \underline{$a_6$} &\underline{$a_7$} & \underline{$a_8$} & \underline{$a_9$}  & \underline{$a_10$}\\
        $ARI$ & 0.11 & 0.11 & 0.11 & 0.1 & 0.09 & 0.1 & 0.09 & 0.09 & 0.1 & 0.09\\
        $\chi^2$ & 0.1 & 0.12 & 0.09 & 0.11 & 0.09 & 0.08 & 0.11 & 0.08 & 0.1 & 0.12\\
        $MI$ & 0.07 & 0.1 & 0.11 & 0.07 & 0.13 & 0.06 & 0.1 & 0.07 & 0.07 & 0.12\\
        $Relief$ & 0.1 & 0.12 & 0.09 & 0.1 & 0.09 & 0.09 & 0.1 & 0.09 & 0.08 & 0.12\\
    $g_5$ & \underline{$a_1$} & \underline{$a_2$} & \underline{$a_3$} &\underline{$a_4$} & \underline{$a_5$} & \underline{$a_6$} &$a_7$ & $a_8$& $a_9$ & $a_{10}$\\
        $ARI$ & 0.18 & 0.18 & 0.16 & 0.16 & 0.16 & 0.15 & 0.0 & 0.0 & 0.0 & 0.0\\
        $\chi^2$ & 0.17 & 0.14 & 0.17 & 0.18 & 0.18 & 0.16 & 0.0 & 0.0 & 0.0 & 0.0\\
        $MI$ & 0.17 & 0.12 & 0.15 & 0.13 & 0.15 & 0.12 & 0.01 & 0.03 & 0.03 & 0.04\\
        $Relief$ & 0.15 & 0.15 & 0.16 & 0.15 & 0.17 & 0.14 & 0.01 & 0.02 & 0.02 & 0.02\\
    $g_6$ & \underline{$a_1$} & \underline{$a_2$} & \underline{$a_3$} & $a_4$& $a_5$&$a_6$ &$a_7$ & $a_8$& $a_9$ & $a_{10}$\\
        $ARI$ & 0.57 & 0.21 & 0.22 & 0.0 & 0.0 & 0.0 & 0.0 & 0.0 & 0.0 & 0.0\\
        $\chi^2$ & 0.81 & 0.07 & 0.11 & 0.0 & 0.0 & 0.0 & 0.0 & 0.0 & 0.0 & 0.0\\
        $MI$ & 0.73 & 0.03 & 0.08 & 0.02 & 0.01 & 0.01 & 0.03 & 0.03 & 0.02 & 0.02\\
        $Relief$ & 0.54 & 0.15 & 0.18 & 0.02 & 0.01 & 0.02 & 0.02 & 0.01 & 0.02 & 0.02\\
    $g_7$ & \underline{$a_1$} & \underline{$a_2$} & \underline{$a_3$} & $a_4$& $a_5$&$a_6$ &$a_7$ & $a_8$& $a_9$ & $a_{10}$\\
        $ARI$ & 0.34 & 0.32 & 0.33 & 0.0 & 0.0 & 0.0 & 0.0 & 0.0 & 0.0 & 0.0\\
        $\chi^2$ & 0.3 & 0.3 & 0.38 & 0.0 & 0.0 & 0.0 & 0.01 & 0.0 & 0.0 & 0.0\\
        $MI$ & 0.23 & 0.13 & 0.28 & 0.05 & 0.03 & 0.04 & 0.03 & 0.08 & 0.03 & 0.02\\
        $Relief$ & 0.3 & 0.3 & 0.35 & 0.01 & 0.0 & 0.01 & 0.01 & 0.0 & 0.01 & 0.01\\
    $g_8$ & \underline{$a_1$} & \underline{$a_2$} & \underline{$a_3$} &\underline{$a_4$} & \underline{$a_5$} & \underline{$a_6$} & \underline{$a_7$}  & \underline{$a_8$}  & \underline{$a_9$}  & \underline{$a_10$} \\
        $ARI$ & 0.1 & 0.1 & 0.1 & 0.1 & 0.11 & 0.09 & 0.1 & 0.09 & 0.1 & 0.12\\
        $\chi^2$ & 0.02 & 0.0 & 0.0 & 0.0 & 0.0 & 0.01 & 0.0 & 0.0 & 0.0 & 0.96\\
        $MI$ & 0.02 & 0.03 & 0.01 & 0.02 & 0.02 & 0.04 & 0.02 & 0.03 & 0.01 & 0.75\\
        $Relief$ & 0.08 & 0.06 & 0.08 & 0.09 & 0.07 & 0.08 & 0.04 & 0.07 & 0.08 & 0.35\\
\end{tabular}
\bigskip
\caption{Comparison $ARI$ - $\chi^2$ - $MI$ - $Relief$ on binary synthetic data - dimension 10  - sample size $1/3$}
\label{arr-chi-mut-artif}
\end{table}
\begin{enumerate}
\item For functions $g_1$ to $g_7$, where the relevant features are identified, $ARI$ provides better results than the other methods:
as soon as $i$ is irrelevant (i.e. does not appear in the function definition), $ARI(i)=0$, while the 3 other methods often provide a non null value.
\item Regarding function $g_8$, it is quite different: $ARI$ provides a larger score for feature $10$, the least significant bit (LSB), which indicates if the number is even. $ARI$ also provides non-negligible scores to the other features. In that perspective, the accuracy of $ARI$ is similar to that of the $Relief$ method. It is clear that $\chi^2$ and $MI$ eliminate all features except the LSB, and it will fail in more than 50\% of the cases.
To give a graphical view of the results from these four methods for function $g_8$, please see the bar charts from Figure \ref{compare-on-g8} in the Appendix.
\end{enumerate}

A discussion regarding the curse of dimensionality will be developed in Subsection \ref{curse}.
\subsection{Extract relevant features on real data: results}
When the relevant features are unknown, which is common in real datasets, the only way to estimate the effectiveness of a given method is to proceed in the following two steps:
\begin{itemize}
    \item Compute the relevant scores with a feature selection method.
    \item Compare the accuracy of using a classifier using all features (baseline) and when using features having the $k$ largest scores as indicated by the feature selection method. The value of $k$ must be the same for all feature selection methods unless a feature has a score of $0$, then this feature has to be eliminated from the list of relevant features. So, it can be the case that we use less than $k$ features in the classifier. We will indicate when we use less than $k$ features.
\end{itemize}
If the accuracy with the reduced set of features is better, it means the target method accurately identifies the relevant features.
Regarding the five UCI target datasets, the relevant features are unknown (except for Monks1). The next sections describe the two-step process.
\subsubsection{Computing scores on real data}
Before comparing the scoring methods' efficiency, let us start by computing the scores with the four different methods. In that case, we follow the same protocol where the sample is $\frac{1}{3}$ of the total available dataset, and we run 10 tests and average the scores per feature and per method.
Results are in Table \ref{air-chi-mut-uci}.
\begin{table}[!ht]
\addtolength{\tabcolsep}{-1.5pt}
\centering
\tiny
\begin{tabular}{c|c|c|c|c|c|c}
$Monks1$ & & & & & &  \\
$ARI$ & 0.37  & 0.38  & 0.0  & 0.0  & 0.25  & 0.0 \\
$\chi^2$ & 0.0  & 0.0  & 0.02  & 0.06  & 0.91  & 0.01 \\
$MI$ & 0.06  & 0.03  & 0.02  & 0.11  & 0.69  & 0.08 \\
$Relief$ & 0.22  & 0.23  & 0.08  & 0.12  & 0.29  & 0.06 
\end{tabular}

\begin{tabular}{c|c|c|c|c|c|c}
$Monks2$ & & & & & &  \\
$ARI$ & 0.15  & 0.14  & 0.23  & 0.14  & 0.11  & 0.23 \\
$\chi^2$ & 0.0  & 0.03  & 0.0  & 0.92  & 0.03  & 0.01 \\
$MI$ & 0.07  & 0.08  & 0.07  & 0.43  & 0.27  & 0.05 \\
$Relief$ & 0.11  & 0.16  & 0.15  & 0.19  & 0.19  & 0.2 
\end{tabular}

\begin{tabular}{c|c|c|c|c|c|c}
$Monks3$ & & & & & &  \\
$ARI$ & 0.03  & 0.46  & 0.02  & 0.07  & 0.4  & 0.02 \\
$\chi^2$ & 0.01  & 0.51  & 0.0  & 0.01  & 0.48  & 0.0 \\
$MI$ & 0.04  & 0.5  & 0.02  & 0.03  & 0.39  & 0.01 \\
$Relief$ & 0.1  & 0.38  & 0.02  & 0.05  & 0.36  & 0.09 
\end{tabular}


\begin{tabular}{c|c|c|c|c|c|c|c|c}
$HIV$  & & & & & & & &  \\
$ARI$ & 0.0  & 0.0  & 0.23  & 0.19  & 0.16  & 0.3  & 0.11  & 0.0 \\
$\chi^2$ & 0.04  & 0.28  & 0.17  & 0.28  & 0.0  & 0.05  & 0.0  & 0.18 \\
$MI$ & 0.13  & 0.06  & 0.06  & 0.23  & 0.22  & 0.1  & 0.08  & 0.12 \\
$Relief$ & 0.12  & 0.14  & 0.11  & 0.16  & 0.13  & 0.11  & 0.14  & 0.11 
\end{tabular}

\begin{tabular}{c|c|c|c|c|c|c|c|c|c|c|c|c|c|c|c|c|c}
$Primary Tumor$ & & & & & & & & & & & & & &   \\
$ARI$ & 0.01 & 0.01 & 0.07 & 0.0 & 0.0 & 0.06 & 0.0 & 0.0 & 0.0 & 0.02 & 0.02 & 0.0 & 0.0 & 0.0 & 0.0 & 0.0 & 0.0\\
$\chi^2$ & 0.67  & 0.0  & 0.09  & 0.04  & 0.04  & 0.0  & 0.0  & 0.03  & 0.0  & 0.01  & 0.05  & 0.0  & 0.0  & 0.01  & 0.02  & 0.0  & 0.01 \\
$MI$ & 0.07  & 0.01  & 0.05  & 0.13  & 0.01  & 0.02  & 0.04  & 0.1  & 0.02  & 0.07  & 0.1  & 0.02  & 0.03  & 0.08  & 0.17  & 0.04  & 0.02 \\
$Relief$ & 0.17  & 0.09  & 0.06  & 0.08  & 0.12  & 0.04  & 0.01  & 0.03  & 0.07  & 0.08  & 0.1  & 0.01  & 0.0  & 0.04  & 0.03  & 0.01  & 0.05 
\end{tabular}
\bigskip
    \caption{Scores from  $ARI$ - $\chi^2$ - $MI$ - $Relief$ on 7 UCI datasets}
    \label{air-chi-mut-uci}
\end{table}
Monks1 is a particular case because the relevant features are known: it is clear that, even with one-third of the dataset, $ARI$ provides perfect scores, when the other methods allocate non-zero scores to known irrelevant features.
\subsubsection{Comparing accuracy with baseline on real data}
According to the above-mentioned process, our final comparison protocol is as follows:
\begin{itemize}
    \item Our baseline classifier is a simple  {\it logistic regression}.
    \item We set the value of $k=4$ and for each method, we classify with the same baseline logistic regression, getting the accuracy as the average value on $10$-folds cross-validation {\bf with a maximum of 4 most relevant features}.
\end{itemize}
The results are in Table \ref{compare-accuracy} where we display the accuracy (in \% rounded to two decimal points). $ARI$ has the highest accuracy in three datasets ($Monks2$, $PrimaryTumor$ and $Mushroom$) - for the $Mushroom$ dataset, the accuracy is higher than when all the features are used. $\chi^2$ also performs better in three datasets ($Monks1$, $BreastCancer$ and $HIV$) and the accuracy for $Monks1$ and $BreastCancer$ are also equally good as $MI$. Note that these are also the two datasets where $MI$ excelled at, and the accuracy
for the $BreastCancer$ dataset is slightly higher compared to when all the features are used. For $Relief$, it only performs best in one dataset ($Monks3$) but it is only slightly higher than $\chi^2$ and $MI$, but much higher than $ARI$. 

Experiment results show that $ARI$ is promising but more work needs to be done to understand why it performs significantly higher or lower than the other three methods in some datasets.

\begin{table}[!ht]
\addtolength{\tabcolsep}{-1.5pt}
    \centering
    \begin{tabular}{c|c|c|c|c|c|c}
    dataset & Baseline & $ARI$ & $\chi^2$ & $MI$ & $Relief$\\
    $Monks1$ & 66.68 & 41.65 & \textbf{66.68} & \textbf{66.68} & 52.57 \\
    $Monks2$ & 63.72 & {\bf 65.72} & 64.22 & 65.39 & 64.56 \\
    $Monks3$ & 76.35 & 49.72 & 76.16 & 75.98 & \textbf{76.53}  \\
    $Breast Cancer$ & 99.29 & 87.34 & \textbf{99.64} & \textbf{99.64} & 52.04  \\
    $HIV$ & 64.20 & 58.05 & {\bf 63.95} & 59.24 & 53.85 \\
    $Primary Tumor$ & 77.31 & \textbf{81.04} & 78.79 & 76.54 & 71.87 \\
    $Mushroom$ & 94.76 & \textbf{93.34} & 81.64 & 86.98 & 70.05 
    \end{tabular}    \bigskip
\caption{Accuracy comparison between $ARI$, $\chi^2$, mutual information and $Relief$ using the 4 most relevant features}
\label{compare-accuracy}
\end{table}
\section{Discussion}\label{discussion}
\subsection{Stability of $ARI$ w.r.t. sample size}
To get an initial validation of the $ARI$ index and check its stability, we only work with binary synthetic datasets of 1024 elements. For each dataset, we sample each dataset with sizes of $100, 200, 400, 500$ and compute $ARI$ for each feature. 
For a fixed size, we average the indices on 10 random samplings. The results are in Table~\ref{valid_feature_relevance_artificial} where the relevant features are underlined.

\begin{table}[!ht]
\addtolength{\tabcolsep}{-1.5pt}
    \centering
    \tiny
    \begin{tabular}{c|c|c|c|c|c|c|c|c|c|c}
    $f$ &  $a_1$ & $a_2$ & $a_3$ & $a_4$ & $a_5$ & $a_6$ & $a_7$ & $a_8$ & $a_9$ & $a_{10}$ \\
    $g_1$ & \underline{$a_1$} & \underline{$a_2$} & \underline{$a_3$} & $a_4$& $a_5$&$a_6$ &$a_7$ & $a_8$& $a_9$ & $a_{10}$\\
        100 & 0.7 & 0.2 & 0.29 & 0.0 & 0.0 & 0.2 & 0.0 & 0.0 & 0.0 & 0.0\\
        200 & 0.73 & 0.27 & 0.32 & 0.0 & 0.0 & 0.0 & 0.0 & 0.0 & 0.0 & 0.0\\
        400 & 0.74 & 0.28 & 0.26 & 0.0 & 0.0 & 0.0 & 0.0 & 0.0 & 0.0 & 0.0\\
        500 & 0.74 & 0.25 & 0.25 & 0.0 & 0.0 & 0.0 & 0.0 & 0.0 & 0.0 & 0.0\\
    $g_2$ & \underline{$a_1$} & \underline{$a_2$} & $a_3$ & $a_4$& $a_5$&$a_6$ &$a_7$ & $a_8$& $a_9$ & $a_{10}$\\
        100 & 1.0 & 1.0 & 0.0 & 0.0 & 0.0 & 0.0 & 0.0 & 0.0 & 0.0 & 0.0\\
        200 & 1.0 & 1.0 & 0.0 & 0.0 & 0.0 & 0.0 & 0.0 & 0.0 & 0.0 & 0.0\\
        400 & 1.0 & 1.0 & 0.0 & 0.0 & 0.0 & 0.0 & 0.0 & 0.0 & 0.0 & 0.0\\
        500 & 1.0 & 1.0 & 0.0 & 0.0 & 0.0 & 0.0 & 0.0 & 0.0 & 0.0 & 0.0\\
    $g_3$ & \underline{$a_1$} & \underline{$a_2$} & \underline{$a_3$} & $a_4$& $a_5$&$a_6$ &$a_7$ & $a_8$& $a_9$ & $a_{10}$\\
        100 & 1.0 & 1.0 & 1.0 & 0.0 & 0.0 & 0.0 & 0.0 & 0.0 & 0.0 & 0.0\\
        200 & 1.0 & 1.0 & 1.0 & 0.0 & 0.0 & 0.0 & 0.0 & 0.0 & 0.0 & 0.0\\
        400 & 1.0 & 1.0 & 1.0 & 0.0 & 0.0 & 0.0 & 0.0 & 0.0 & 0.0 & 0.0\\
        500 & 1.0 & 1.0 & 1.0 & 0.0 & 0.0 & 0.0 & 0.0 & 0.0 & 0.0 & 0.0\\
    $g_4$ & \underline{$a_1$} & \underline{$a_2$} & \underline{$a_3$} &\underline{$a_4$} & \underline{$a_5$} & \underline{$a_6$} &\underline{$a_7$} & \underline{$a_8$} & \underline{$a_9$}  & \underline{$a_10$}\\
        100 & 0.25 & 0.22 & 0.28 & 0.21 & 0.2 & 0.19 & 0.19 & 0.27 & 0.29 & 0.27\\
        200 & 0.2 & 0.19 & 0.23 & 0.21 & 0.21 & 0.18 & 0.23 & 0.21 & 0.19 & 0.18\\
        400 & 0.22 & 0.23 & 0.22 & 0.2 & 0.21 & 0.2 & 0.23 & 0.19 & 0.22 & 0.21\\
        500 & 0.22 & 0.21 & 0.25 & 0.24 & 0.24 & 0.23 & 0.24 & 0.23 & 0.23 & 0.23\\ 
    $g_5$ & \underline{$a_1$} & \underline{$a_2$} & \underline{$a_3$} &\underline{$a_4$} & \underline{$a_5$} & \underline{$a_6$} &$a_7$ & $a_8$& $a_9$ & $a_{10}$\\
        100 & 0.2 & 0.28 & 0.26 & 0.44 & 0.23 & 0.16 & 0.0 & 0.0 & 0.0 & 0.0\\
        200 & 0.24 & 0.28 & 0.21 & 0.18 & 0.18 & 0.25 & 0.0 & 0.0 & 0.0 & 0.0\\
        400 & 0.24 & 0.23 & 0.24 & 0.22 & 0.22 & 0.22 & 0.0 & 0.0 & 0.0 & 0.0\\
        500 & 0.22 & 0.24 & 0.22 & 0.21 & 0.22 & 0.2 & 0.0 & 0.0 & 0.0 & 0.0\\
    $g_6$ & \underline{$a_1$} & \underline{$a_2$} & \underline{$a_3$} & $a_4$& $a_5$&$a_6$ &$a_7$ & $a_8$& $a_9$ & $a_{10}$\\
        100 & 0.78 & 0.19 & 0.16 & 0.0 & 0.0 & 0.0 & 0.0 & 0.0 & 0.0 & 0.0\\
        200 & 0.74 & 0.23 & 0.26 & 0.0 & 0.0 & 0.0 & 0.0 & 0.0 & 0.0 & 0.0\\
        400 & 0.74 & 0.25 & 0.25 & 0.0 & 0.0 & 0.0 & 0.0 & 0.0 & 0.0 & 0.0\\
        500 & 0.77 & 0.25 & 0.24 & 0.0 & 0.0 & 0.0 & 0.0 & 0.0 & 0.0 & 0.0\\
    $g_7$ & \underline{$a_1$} & \underline{$a_2$} & \underline{$a_3$} & $a_4$& $a_5$&$a_6$ &$a_7$ & $a_8$& $a_9$ & $a_{10}$\\
        100 & 0.75 & 0.67 & 0.8 & 0.0 & 0.0 & 0.0 & 0.0 & 0.0 & 0.0 & 0.0\\
        200 & 0.7 & 0.74 & 0.76 & 0.0 & 0.0 & 0.0 & 0.0 & 0.0 & 0.0 & 0.0\\
        400 & 0.75 & 0.76 & 0.75 & 0.0 & 0.0 & 0.0 & 0.0 & 0.0 & 0.0 & 0.0\\
        500 & 0.74 & 0.73 & 0.75 & 0.0 & 0.0 & 0.0 & 0.0 & 0.0 & 0.0 & 0.0\\
    $g_8$ & \underline{$a_1$} & \underline{$a_2$} & \underline{$a_3$} &\underline{$a_4$} & \underline{$a_5$} & \underline{$a_6$} & \underline{$a_7$}  & \underline{$a_8$}  & \underline{$a_9$}  & \underline{$a_10$} \\
        100 & 0.31 & 0.26 & 0.17 & 0.22 & 0.25 & 0.29 & 0.12 & 0.18 & 0.33 & 0.45\\
        200 & 0.27 & 0.23 & 0.26 & 0.29 & 0.27 & 0.22 & 0.27 & 0.3 & 0.28 & 0.31\\
        400 & 0.25 & 0.25 & 0.24 & 0.28 & 0.24 & 0.23 & 0.24 & 0.24 & 0.27 & 0.33\\
        500 & 0.28 & 0.26 & 0.27 & 0.3 & 0.27 & 0.25 & 0.25 & 0.26 & 0.27 & 0.35\\
    \end{tabular}
    \bigskip
    \caption{Initial validation for $ARI$}
    \label{valid_feature_relevance_artificial}
\end{table}
At this stage, the scores are not normalized and their sum can be greater than $1$. As can be seen in Table \ref{valid_feature_relevance_artificial}: 
\begin{itemize}
    \item Whatever the sample size, the order of the values of $ARI$ remains stable for a given dataset: the highest values of $ari$ are obtained for the most relevant features when they are known. 
    \item When sample size is larger than $200$, irrelevant features $i$ are all pointed out with $ARI(i)=0$.
    \item As expected, our indices do not provide any information when all features are relevant as is the case for functions
    $g_5$ and $g_8$.
    None of the values of $ARI$ are $1$ or $0$.
    \item It is interesting to note that, for $g_8$ function (prime number example), $ARI$ does not provide any hint about the particular relevance of a given digit, each one having an index between $0.25$ and $0.30$, except the last one (corresponding to the Least Significant Bit). In fact, the LSB holds a strong information content regarding primality: if LSB=$0$ then the number is not prime except when it is $2$.
\end{itemize}
\subsection{Curse of dimensionality and categorical range}\label{curse}
Being a statistical test, $ARI$ is sensitive to the curse of dimensionality, and more generally to the size of the sample with regard to the total size of the universe $X$. 
Roughly speaking, for a fixed sample size, the significance of the test decreases when the size of $X$ increases.
This happens when the dimensionality of $X$ increases or when
the ranges $X_i$ of categorical features increase.
Let us for instance consider the following table where the sample size is fixed to 500 ( $range=2$ means binary features):
\begin{table}[ht!]
    \centering
    \begin{tabular}{c|c|r|c|r}
    \hline
    dimension & range & cardinality of $X$ & sample size $S$ & average percentage of $S$ over $X$\\
    \hline
         10 & 2 & 1024 & 500 & 48.8\% \\
         \hline
         10 & 3 & 59 049 & 500 & 0.8\% \\
         \hline
         15 & 2 & 32 768 & 500 & 1.5\% \\
         \hline
         15 & 3 & 14 348 907 & 500 & 0.00004\%\\
         \hline
         15 & 3 & 14 348 907 & 10,00 & 0.00008\%\\
         \hline
         15 & 3 & 14 348 907 & 10,000 & 0.001\%\\
         \hline
    \end{tabular}
    \bigskip
    \caption{Curse of dimensionality and range of features.}
    \label{dimensionality-range}
\end{table}
Computing $ARI(i)$ leads to looking for pairs $(a,b)$ differing only on feature $i$. 
When the dimensionality and/or the range of features become high then $X$ is large. If the sample size $S$ is small, then it is very unlikely to have in $S$, 2 elements differing only on $i$ i.e. any Hamming ball of radius $1$ is just empty.
As a consequence, $Dif(i)^S$ is likely to be empty and, if $i$ has a non-zero variance, then $ARI(i)=2$. 

We then experiment on artificial data, with the same functions $g_1,\ldots,g_8$, but where we simply extend the dimension to 15 and categorical range to $\{0,1,2\}$ (i.e. $|X| > 14 000 000$). When the size of $S$ is relatively small, whatever the feature $i$, $Dif(i)^S$ is likely empty and we get $ARI(i)=2$, which does not provide any relevant information at this stage. 
In the annex, Table \ref{dimensionality} gives an overview of what size is needed for $S$ to get relevant values from $ARI$.
With a sample size of 5000, $ARI$ starts to discriminate between features. When moving to samples of 10,000 elements (i.e. 0.0008\% of the total size of $X$), $ARI$ provides accurate information regarding relevance/irrelevance.  Obviously, more experiments have to be carried out to get a better understanding of the impact of dimensionality and range on $ARI$. Also, other options to overcome the issues have to be investigated.

\section{Conclusion}\label{conc}
We have developed a new index to estimate feature relevance/irrelevance. Inspired by the concept of analogical proportion, we compare 2 elements of a given sample, which do not have the same label and differ only on one feature (their Hamming distance is $1$). Such a pair of elements is a marker of the relevance of the target feature to the label. 
The more we have such pairs, the more relevant is the target feature.
This leads us to compute the Analogical Relevance Ratio $ARI$.
Preliminary experiments highlight the power of $ARI$ both on artificial and real datasets.
Being a purely statistical index, $ARI$ is obviously subject to the curse of dimensionality. 
It means that, when dimension or value range increases, it is quite unlikely to get relevant $ARI$ values.
Nevertheless, a track to overcome this issue is to consider a relaxed definition of $ARI$ where we do not limit the distance between 2 elements to be exactly $1$. 
At this stage, $ARI$ is dedicated to categorical features. We could also take inspiration from the $Relief$ method, to extend $ARI$ to real-valued features.
\bibliographystyle{unsrtnat}
\bibliography{biblio}

\newpage
\appendix

\section{Bar chart for comparing $ARI$, $\chi^2$, $MI$ and $Relief$ on function $g_8$}
\begin{figure}[!ht]
\tiny
    \centering
    \begin{minipage}{0.45\linewidth}
      \includegraphics[scale=0.5]{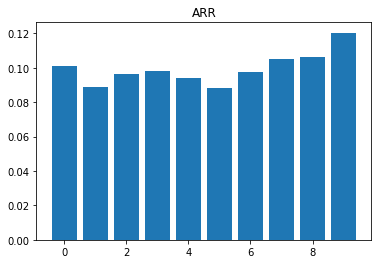}
      \caption{$ARI$}
    \end{minipage}
    \hfill
    \begin{minipage}{0.45\linewidth}
      \includegraphics[scale=0.5]{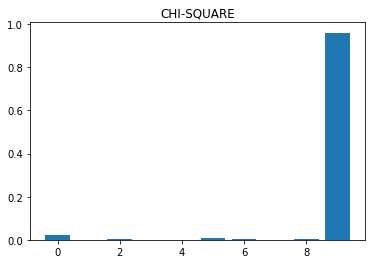}
      \caption{$\chi^2$}
    \end{minipage}

    \centering
    \begin{minipage}{0.45\linewidth}
      \includegraphics[scale=0.5]{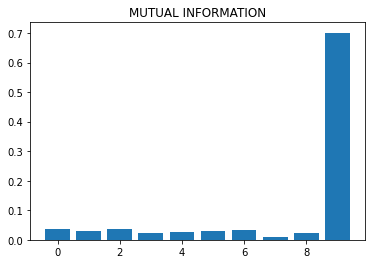}
      \caption{$MI$}
    \end{minipage}
    \hfill
    \begin{minipage}{0.45\linewidth}
      \includegraphics[scale=0.5]{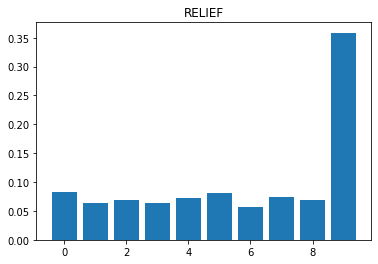}
      \caption{$Relief$}
    \end{minipage}
    \bigskip
    \caption{Comparison on g8 function}
    \label{compare-on-g8}
\end{figure}

We give a graphic view of what the 4 methods compute when function $g_8$ provides the label: 
the bar chart in Figure \ref{compare-on-g8} shows that except $ARI$, all methods give the Least Significant Bit (LSB) a very important score.
Especially $\chi^2$ where all the other bits are considered as more or less not significant.
But, in fact, when the LSB is equal to $1$, the remaining digits are significant. In that perspective, $ARI$ seems close to $Relief$.
\newpage
\section{Curse of dimensionality and range}
We experiment on a synthetic dataset of dimension $15$, with a categorical range of $3$ (i.e. each feature can take 3 different values).
The theoretical size of the whole universe $X$ is then larger than $14 000 000$. We average $ARI$ scores on 10 samples of fixed size, the size being $400, 500, 1000, 2000, 5000$ and $10000$.
 \begin{table}[ht!]
     \centering \tiny
     \begin{tabular}{c|c|c|c|c|c|c|c|c|c|c|c|c|c|c|c}
     \hline \\
     $g_1$ & $a_1$ & $a_2$ & $a_3$ & $a_4$ & $a_5$ & $a_6$ & $a_7$ & $a_8$ & $a_9$ & $a_{10}$ & $a_{11}$ & $a_{12}$ & $a_{13}$ & $a_{14}$ & $a_{15}$ \\
     \hline \\
 400 & 2  & 2  & 2  & 2  & 2  & 2  & 2  & 2  & 1.8-0.6 & 2  & 2  & 2  & 1.8-0.6 & 2  & 2 \\
 500 & 2  & 2  & 2  & 2  & 2  & 2  & 2  & 2  & 2  & 2  & 2  & 2  & 1.8-0.6 & 2  & 2 \\
 1000 & 2  & 1.8-0.6 & 2  & 2  & 2  & 2  & 1.8-0.6 & 2  & 2  & 2  & 2  & 2  & 2  & 1.8-0.6 & 1.8-0.6\\
 2000 & 2  & 1.8-0.6 & 1.3-0.9 & 0.8-0.98 & 1.6-0.8 & 1.4-0.92 & 1.2-0.98 & 1.6-0.8 & 1.6-0.8 & 1.2-0.98 & 1.4-0.92 & 2  & 1.8-0.6 & 1.8-0.6 & 1.2-0.98\\
 5000 & 0.78-0.69 & 0.1-0.2 & 0.56-0.78 & 0.2-0.6 & 0.4-0.8 & 0.2-0.6 & 0.4-0.8 & 0.2-0.6 & 0.6-0.92 & 0.2-0.6 & 0.2-0.6 & 0.6-0.92 & 0.6-0.92 & 0.4-0.8 & 0.4-0.8\\
 10000 & 0.61-0.23 & 0.15-0.14 & 0.17-0.11 & 0.0  & 0.0  & 0.0  & 0.0  & 0.0  & 0.0  & 0.0  & 0.0  & 0.0  & 0.0  & 0.0  & 0.0 \\
 \hline \\\hline \\
 \hline \\
     $g_2$ & $a_1$ & $a_2$ & $a_3$ & $a_4$ & $a_5$ & $a_6$ & $a_7$ & $a_8$ & $a_9$ & $a_{10}$ & $a_{11}$ & $a_{12}$ & $a_{13}$ & $a_{14}$ & $a_{15}$ \\
     \hline \\
400 & 2  & 2  & 2  & 2  & 2  & 2  & 2  & 2  & 2  & 2  & 2  & 2  & 1.6-0.8 & 2  & 2 \\
500 & 1.8-0.6 & 1.9-0.3 & 2  & 2  & 1.8-0.6 & 2  & 2  & 2  & 2  & 2  & 2  & 2  & 1.8-0.6 & 2  & 2 \\
1000 & 2  & 1.8-0.6 & 1.6-0.8 & 2  & 1.8-0.6 & 2  & 2  & 2  & 2  & 1.6-0.8 & 2  & 1.8-0.6 & 2  & 1.8-0.6 & 2 \\
2000 & 1.9-0.3 & 1.4-0.92 & 2  & 1.6-0.8 & 1.6-0.8 & 1.2-0.98 & 1.6-0.8 & 1.6-0.8 & 1.6-0.8 & 1.8-0.6 & 1.6-0.8 & 1.4-0.92 & 1.4-0.92 & 1.6-0.8 & 1.2-0.98\\
5000 & 1.14-0.44 & 0.93-0.77 & 0.0  & 0.6-0.92 & 0.2-0.6 & 0.2-0.6 & 0.0  & 0.4-0.8 & 0.2-0.6 & 0.2-0.6 & 0.2-0.6 & 0.0  & 0.0  & 0.4-0.8 & 0.6-0.92\\
10000 & 0.67-0.16 & 0.59-0.15 & 0.0  & 0.0  & 0.0  & 0.0  & 0.0  & 0.0  & 0.0  & 0.0  & 0.0  & 0.0  & 0.0  & 0.0  & 0.0 \\
 \hline \\\hline \\
 \hline \\
     $g_3$ & $a_1$ & $a_2$ & $a_3$ & $a_4$ & $a_5$ & $a_6$ & $a_7$ & $a_8$ & $a_9$ & $a_{10}$ & $a_{11}$ & $a_{12}$ & $a_{13}$ & $a_{14}$ & $a_{15}$ \\
     \hline \\
400 & 2  & 2  & 2  & 2  & 2  & 2  & 2  & 2  & 2  & 2  & 2  & 1.8-0.6 & 2  & 2  & 2 \\
500 & 2  & 2  & 1.9-0.3 & 2  & 2  & 2  & 2  & 2  & 1.8-0.6 & 2  & 2  & 2  & 2  & 2  & 2 \\
1000 & 1.9-0.3 & 1.9-0.3 & 2  & 1.8-0.6 & 1.8-0.6 & 2  & 2  & 1.8-0.6 & 2  & 1.8-0.6 & 1.8-0.6 & 1.8-0.6 & 2  & 2  & 2 \\
2000 & 1.6-0.66 & 1.6-0.66 & 1.5-0.67 & 1.0-1.0 & 1.0-1.0 & 1.8-0.6 & 1.6-0.8 & 1.4-0.92 & 1.8-0.6 & 1.2-0.98 & 1.6-0.8 & 1.4-0.92 & 1.4-0.92 & 1.4-0.92 & 1.8-0.6\\
5000 & 1.05-0.68 & 0.82-0.55 & 1.09-0.66 & 0.6-0.92 & 1.0-1.0 & 0.2-0.6 & 0.2-0.6 & 0.2-0.6 & 0.4-0.8 & 0.4-0.8 & 0.2-0.6 & 0.4-0.8 & 0.6-0.92 & 0.2-0.6 & 0.2-0.6\\
10000 & 0.72-0.18 & 0.59-0.24 & 0.65-0.19 & 0.0  & 0.0  & 0.0  & 0.0  & 0.0  & 0.0  & 0.0  & 0.0  & 0.0  & 0.0  & 0.0  & 0.0 \\
 \hline \\\hline \\
 \hline \\
     $g_4$ & $a_1$ & $a_2$ & $a_3$ & $a_4$ & $a_5$ & $a_6$ & $a_7$ & $a_8$ & $a_9$ & $a_{10}$ & $a_{11}$ & $a_{12}$ & $a_{13}$ & $a_{14}$ & $a_{15}$ \\
     \hline \\
 400 & 2  & 2  & 2  & 2  & 2  & 2  & 2  & 2  & 2  & 2  & 2  & 1.8-0.6 & 2  & 1.8-0.6 & 1.8-0.6\\
 500 & 1.8-0.6 & 2  & 2  & 2  & 2  & 2  & 2  & 2  & 2  & 2  & 2  & 2  & 1.8-0.6 & 2  & 2 \\
 1000 & 2  & 2  & 1.6-0.8 & 1.8-0.6 & 2  & 2  & 1.8-0.6 & 2  & 1.6-0.8 & 1.6-0.8 & 2  & 1.8-0.6 & 2  & 1.8-0.6 & 1.8-0.6\\
 2000 & 1.6-0.8 & 1.0-1.0 & 1.2-0.98 & 1.4-0.92 & 1.4-0.92 & 1.4-0.92 & 1.4-0.92 & 2  & 1.8-0.6 & 1.4-0.92 & 1.2-0.98 & 1.4-0.92 & 1.4-0.92 & 1.6-0.8 & 1.6-0.8\\
 5000 & 0.2-0.6 & 0.2-0.6 & 0.4-0.8 & 0.4-0.8 & 0.4-0.8 & 0.2-0.6 & 0.2-0.6 & 0.0  & 0.6-0.92 & 0.2-0.6 & 0.2-0.6 & 0.2-0.6 & 0.6-0.92 & 0.6-0.92 & 0.2-0.6\\
 10000 & 0.0  & 0.0  & 0.0  & 0.0  & 0.0  & 0.0  & 0.2-0.6 & 0.0  & 0.0  & 0.0  & 0.0  & 0.0  & 0.0  & 0.0  & 0.0 \\
 \hline \\\hline \\\hline \\
     $g_5$ & $a_1$ & $a_2$ & $a_3$ & $a_4$ & $a_5$ & $a_6$ & $a_7$ & $a_8$ & $a_9$ & $a_{10}$ & $a_{11}$ & $a_{12}$ & $a_{13}$ & $a_{14}$ & $a_{15}$ \\
     \hline \\
 400 & 2  & 2  & 2  & 1.6-0.8 & 2  & 2  & 1.8-0.6 & 2  & 2  & 2  & 2  & 1.8-0.6 & 2  & 2  & 2 \\
 500 & 2  & 2  & 1.8-0.6 & 2  & 2  & 1.8-0.6 & 2  & 2  & 2  & 1.8-0.6 & 2  & 2  & 2  & 2  & 2 \\
 1000 & 2  & 2  & 1.8-0.6 & 2  & 2  & 2  & 1.6-0.8 & 1.6-0.8 & 1.6-0.8 & 1.8-0.6 & 2  & 1.8-0.6 & 2  & 1.8-0.6 & 1.8-0.6\\
 2000 & 1.6-0.8 & 1.7-0.64 & 1.0-1.0 & 1.0-1.0 & 1.5-0.81 & 2  & 1.2-0.98 & 1.8-0.6 & 1.2-0.98 & 1.6-0.8 & 1.2-0.98 & 2  & 1.6-0.8 & 1.0-1.0 & 1.4-0.92\\
 5000 & 0.22-0.6 & 0.62-0.9 & 0.8-0.98 & 0.52-0.76 & 0.2-0.6 & 0.55-0.79 & 0.8-0.98 & 0.2-0.6 & 0.2-0.6 & 0.2-0.6 & 0.0  & 0.2-0.6 & 0.6-0.92 & 0.4-0.8 & 0.8-0.98\\
 10000 & 0.09-0.1 & 0.02 4 & 0.08-0.1 & 0.17-0.19 & 0.02 5 & 0.08-0.12 & 0.0  & 0.0  & 0.0  & 0.0  & 0.0  & 0.0  & 0.0  & 0.0  & 0.0 \\
 \hline \\ \hline
 $g_6$ & $a_1$ & $a_2$ & $a_3$ & $a_4$ & $a_5$ & $a_6$ & $a_7$ & $a_8$ & $a_9$ & $a_{10}$ & $a_{11}$ & $a_{12}$ & $a_{13}$ & $a_{14}$ & $a_{15}$ \\
     \hline \\
  400 & 2  & 2  & 2  & 2  & 1.8-0.6 & 2  & 2  & 2  & 2  & 2  & 2  & 2  & 1.8-0.6 & 2  & 2 \\
 500 & 2  & 2  & 2  & 2  & 2  & 2  & 2  & 2  & 2  & 1.8-0.6 & 2  & 2  & 2  & 2  & 1.8-0.6\\
 1000 & 1.8-0.6 & 1.8-0.6 & 1.6-0.8 & 2  & 1.4-0.92 & 2  & 2  & 2  & 2  & 1.6-0.8 & 2  & 1.8-0.6 & 1.8-0.6 & 1.6-0.8 & 1.8-0.6\\
 2000 & 1.8-0.6 & 1.8-0.4 & 1.4-0.92 & 1.6-0.8 & 1.4-0.92 & 1.6-0.8 & 1.4-0.92 & 1.6-0.8 & 1.0-1.0 & 1.4-0.92 & 1.4-0.92 & 1.8-0.6 & 1.2-0.98 & 2  & 1.0-1.0\\
 5000 & 0.84-0.68 & 0.88-0.78 & 0.62-0.75 & 0.4-0.8 & 0.4-0.8 & 0.6-0.92 & 0.8-0.98 & 0.6-0.92 & 0.6-0.92 & 0.0  & 0.2-0.6 & 0.0  & 0.6-0.92 & 0.2-0.6 & 0.4-0.8\\
 10000 & 0.47-0.27 & 0.28-0.16 & 0.16-0.12 & 0.0  & 0.0  & 0.0  & 0.0  & 0.0  & 0.0  & 0.0  & 0.0  & 0.0  & 0.0  & 0.0  & 0.0 \\
 \hline \\\hline \\
     $g_7$ & $a_1$ & $a_2$ & $a_3$ & $a_4$ & $a_5$ & $a_6$ & $a_7$ & $a_8$ & $a_9$ & $a_{10}$ & $a_{11}$ & $a_{12}$ & $a_{13}$ & $a_{14}$ & $a_{15}$ \\
     \hline \\
 300 & 2  & 2  & 2  & 2  & 2  & 2  & 1.8-0.6 & 2  & 2  & 2  & 2  & 2  & 2  & 2  & 2 \\
 400 & 2  & 2  & 2  & 1.8-0.6 & 2  & 2  & 2  & 2  & 2  & 2  & 2  & 2  & 2  & 2  & 2 \\
 500 & 2  & 1.9-0.3 & 2  & 2  & 2  & 2  & 2  & 2  & 2  & 1.8-0.6 & 2  & 2  & 2  & 2  & 2 \\
 1000 & 2  & 2  & 2  & 1.6-0.8 & 2  & 2  & 1.8-0.6 & 1.8-0.6 & 2  & 2  & 2  & 2  & 2  & 2  & 2 \\
 2000 & 1.5-0.67 & 1.6-0.66 & 1.8-0.6 & 1.8-0.6 & 1.6-0.8 & 1.2-0.98 & 1.6-0.8 & 1.8-0.6 & 1.6-0.8 & 1.4-0.92 & 1.2-0.98 & 1.8-0.6 & 1.6-0.8 & 1.2-0.98 & 1.6-0.8\\
 5000 & 0.45-0.61 & 0.45-0.36 & 0.89-0.64 & 0.6-0.92 & 0.0  & 0.8-0.98 & 0.4-0.8 & 0.8-0.98 & 0.2-0.6 & 0.2-0.6 & 0.6-0.92 & 0.2-0.6 & 0.6-0.92 & 0.6-0.92 & 0.0 \\
 10000 & 0.46-0.2 & 0.26-0.21 & 0.53-0.22 & 0.0  & 0.0  & 0.0  & 0.0  & 0.0  & 0.0  & 0.0  & 0.0  & 0.0  & 0.0  & 0.0  & 0.0 \\
 \hline \\\hline \\
     $g_8$ & $a_1$ & $a_2$ & $a_3$ & $a_4$ & $a_5$ & $a_6$ & $a_7$ & $a_8$ & $a_9$ & $a_{10}$ & $a_{11}$ & $a_{12}$ & $a_{13}$ & $a_{14}$ & $a_{15}$ \\
     \hline \\
 400 & 2  & 2  & 2  & 2  & 2  & 2  & 2  & 2  & 1.8-0.6 & 2  & 2  & 2  & 1.8-0.6 & 2  & 2 \\
 500 & 2  & 2  & 2  & 2  & 2  & 2  & 2  & 2  & 2  & 2  & 2  & 2  & 1.8-0.6 & 2  & 2 \\
 1000 & 2  & 2  & 2  & 2  & 1.8-0.6 & 1.6-0.8 & 2  & 1.6-0.8 & 1.6-0.8 & 1.8-0.6 & 2  & 2  & 2  & 2  & 2 \\
 2000 & 1.4-0.92 & 1.6-0.8 & 2  & 1.4-0.92 & 1.45-0.85 & 1.4-0.92 & 1.4-0.92 & 1.8-0.6 & 1.4-0.92 & 1.4-0.92 & 1.2-0.98 & 1.0-0.89 & 1.8-0.6 & 1.8-0.6 & 1.6-0.66\\
 5000 & 0.8-0.98 & 1.08-0.93 & 0.28-0.59 & 0.6-0.92 & 0.83-0.96 & 0.15-0.23 & 0.87-0.93 & 0.67-0.88 & 0.65-0.9 & 0.1-0.3 & 0.45-0.79 & 0.57-0.74 & 0.3-0.6 & 0.53-0.79 & 0.35-0.63\\
 10000 & 0.61-0.23 & 0.15-0.14 & 0.17-0.11 & 0.0  & 0.0  & 0.0  & 0.0  & 0.0  & 0.0  & 0.0  & 0.0  & 0.0  & 0.0  & 0.0  & 0.0 \\
 \hline
     \end{tabular}
     \bigskip
     \caption{$ARI$ values from synthetic data sets using different sample sizes.}
     \label{dimensionality}
 \end{table}

\end{document}